\documentclass[11pt]{article}
\usepackage{ACL2023}

\usepackage{times}
\usepackage{latexsym}

\usepackage{colortbl}

\usepackage{graphicx}
\usepackage{paralist}
\usepackage{amsmath}
\usepackage[most]{tcolorbox}
\usepackage[linesnumbered,ruled,vlined]{algorithm2e}

\usepackage{amssymb}
\usepackage{amsmath}
\usepackage[ruled,vlined]{algorithm2e}

\usepackage[T1]{fontenc}

\usepackage[utf8]{inputenc}

\usepackage{microtype}

\usepackage{inconsolata}

\usepackage{booktabs}     
\usepackage{booktabs}
\usepackage{multirow}

\definecolor{nmgray}{RGB}{229,229,229}

\usepackage{graphicx}

\newtcolorbox{mybox}[2][]{
width=\columnwidth,
colback = nmgray!75!white, 
colframe = nmgray!75!white, 
boxsep=0pt,left=9pt,right=10pt,top=0pt,bottom=0pt,
fontupper=\linespread{0.9}\selectfont,
title=#2,#1}

\usepackage{hyperref}

\newcommand{\translationLogo}{\raisebox{-3pt}{\includegraphics[width=1.2em]{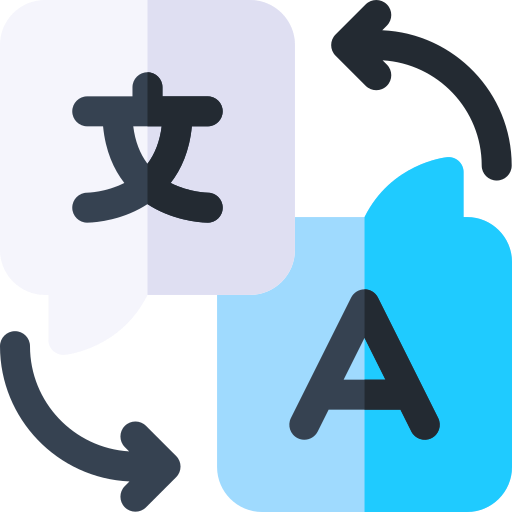}}\xspace}

\newcommand{\decomposerLogo}{\raisebox{-2.8pt}{\includegraphics[width=1.25em]{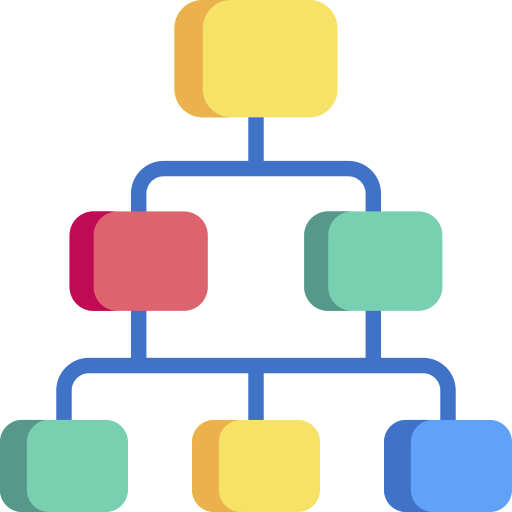}}\xspace}

\newcommand{\searcherLogo}{\raisebox{-3pt}{\includegraphics[width=1.25em]{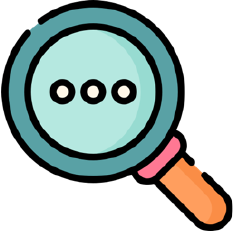}}\xspace}

\newcommand{\resolverLogo}{\raisebox{-2.7pt}{\includegraphics[width=1.25em]{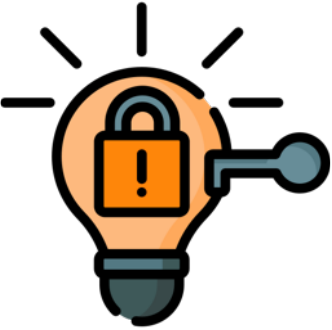}}\xspace}

\newcommand{\aristotleLogo}{\raisebox{-2.7pt}{\includegraphics[width=1em]{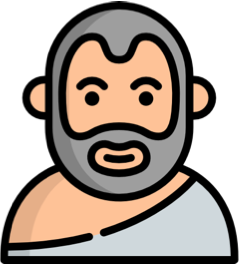}}\xspace}

\title{\texttt{\aristotleLogo Aristotle}: Mastering Logical Reasoning with\\ A Logic-Complete Decompose-Search-Resolve Framework}

\author{
\textbf{Jundong Xu}\textsuperscript{\rm 1}, \quad
\textbf{Hao Fei}\textsuperscript{\rm 1}\thanks{\, Corresponding author: Hao Fei}, \quad
\textbf{Meng Luo}\textsuperscript{\rm 1}, \quad
\textbf{Qian Liu}\textsuperscript{\rm 2}, \quad
\textbf{Liangming Pan}\textsuperscript{\rm 3}, \\[0.3ex]
\textbf{William Yang Wang}\textsuperscript{\rm 4}, \quad
\textbf{Preslav Nakov}\textsuperscript{\rm 5}, \quad
\textbf{Mong-Li Lee}\textsuperscript{\rm 1}, \quad
\textbf{Wynne Hsu}\textsuperscript{\rm 1} \\
\textsuperscript{\rm 1} National University of Singapore, 
\textsuperscript{\rm 2} University of Auckland, 
\textsuperscript{\rm 3} University of Arizona, \\
\textsuperscript{\rm 4} University of California, Santa Barbara \quad
\textsuperscript{\rm 5} MBZUAI \\
\tt{jundong.xu@u.nus.edu\quad haofei37@nus.edu.sg\quad mluo@u.nus.edu} \\
\tt{liu.qian@auckland.ac.nz\quad liangmingpan@arizona.edu\quad william@cs.ucsb.edu} \\
\tt{preslav.nakov@mbzuai.ac.ae\quad dcsleeml@nus.edu.sg\quad dcshsuw@nus.edu.sg}
}

\begin{document}
\maketitle
\begin{abstract}
In the context of large language models (LLMs), current advanced reasoning methods
have made impressive strides in various reasoning tasks.  
However, when it comes to \emph{logical} reasoning tasks, major challenges remain in both efficacy and efficiency.
This is rooted in the fact that these systems fail to fully leverage the inherent structure of logical tasks throughout the reasoning processes such as decomposition, search, and resolution.  
To address this, we propose a logic-complete reasoning framework, \textbf{\texttt{Aristotle}},  
with three key components: \emph{Logical Decomposer}, \emph{Logical Search Router}, and \emph{Logical Resolver}. In our framework, symbolic expressions and logical rules are comprehensively integrated into the entire reasoning process, significantly alleviating the bottlenecks of logical reasoning, i.e., reducing sub-task complexity, minimizing search errors, and resolving logical contradictions.  
The experimental results on several datasets demonstrate that \texttt{Aristotle} consistently outperforms state-of-the-art reasoning frameworks in both accuracy and efficiency, particularly excelling in complex logical reasoning scenarios.
We will open-source all our code at \url{http://llm-symbol.github.io/Aristotle}.
\end{abstract}

\section{Introduction}

LLMs \cite{gpt3, palm} have unlocked unprecedented potential in semantic understanding \cite{langauge-understanding}, sparking immense hope for realizing AGI.
A fundamental requirement for true intelligence is the ability to perform human-level reasoning, such as commonsense reasoning \cite{commonsense-reason}, mathematical problem-solving \cite{math-reason}, and geometric reasoning \cite{geometric-reason}.
To achieve this, researchers have drawn inspiration from human reasoning processes, proposing various methods and strategies for LLM-based reasoning.
One of the most groundbreaking works is the Chain-of-Thought (CoT) \cite{cot}, which breaks down complex problems into smaller sub-problems, solving them step by step.
The birth of CoT has elevated the reasoning capabilities of LLMs to new heights.
Further research has built on this foundation by closely emulating human cognitive patterns, introducing more advanced approaches, such as Least-to-Most \cite{least-to-most}, Tree-of-Thought (ToT) \cite{ToT}, Graph-of-Thought (GoT) \cite{got}, and Plan-and-Solve \cite{plan-and-solve}, which have achieved progressively better results on reasoning benchmarks.
In summary, successful LLM-based reasoning methods generally involve three key modules \cite{logic-reasoning, reason-plan-survey}: \textbf{problem decomposition}, \textbf{path searching}, and \textbf{problem resolution}.

\vspace{-1mm}
\begin{figure}[!t]
\centering
\resizebox{0.98\columnwidth}{!}{
  \includegraphics{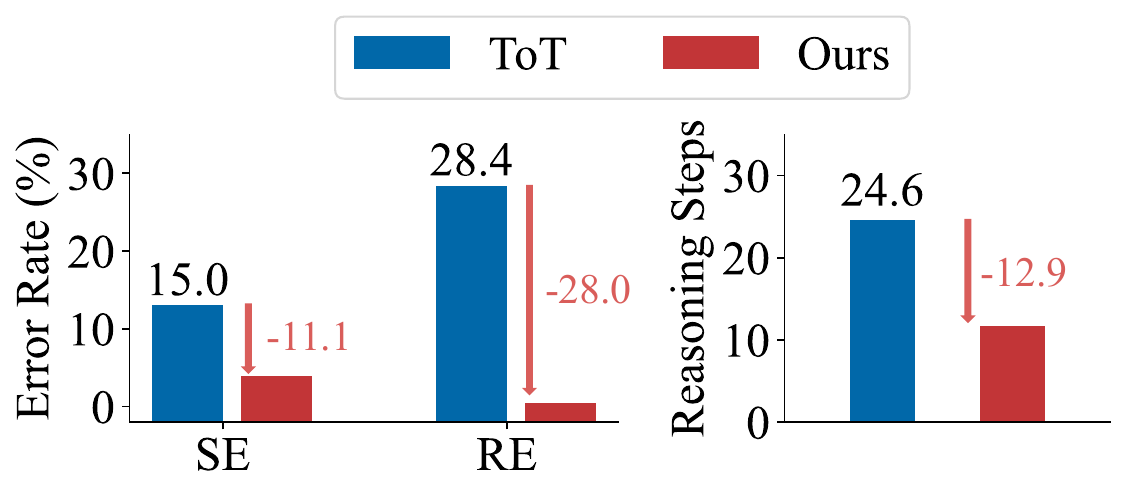}
}
\vspace{-2mm}
\caption{\textbf{Our reasoning framework vs. the SoTA ToT:} comparison in terms of Search Error (SE) and single-step Reasoning Error (RE), as well as in terms of average number of reasoning steps.
}
\label{intro}
\vspace{-4mm}
\end{figure}

\begin{figure*}[!t]
\centering
\resizebox{\textwidth}{!}{
  \includegraphics{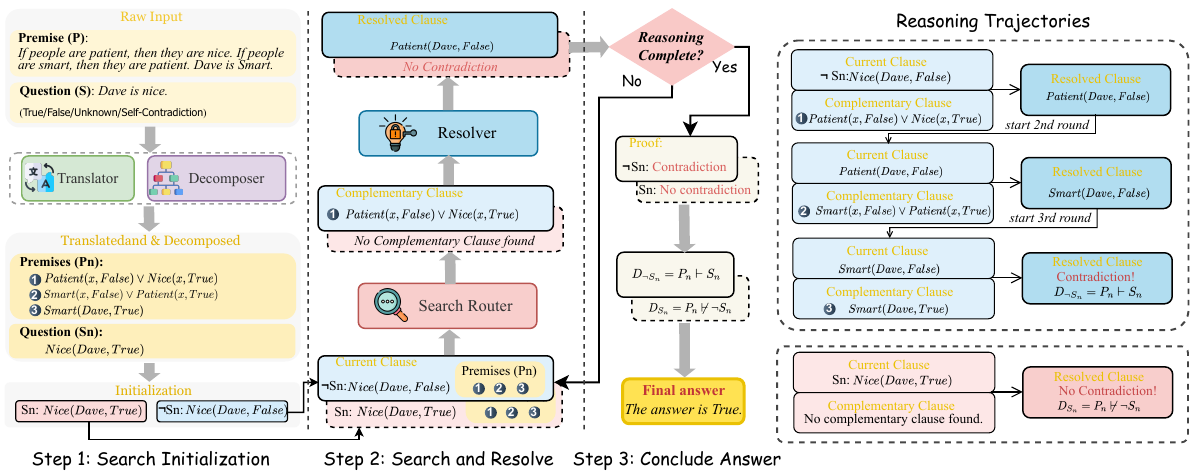}
}
\caption{Our \texttt{Aristotle} logical reasoning framework (best viewed via zooming in).
In \textbf{step 1}, the \translationLogo Translator and \decomposerLogo Decomposer together transform \(P\) and \(S\) into \(P_n\) and \(S_n\). 
Then, we initialize the \(C_{\text{current}}\) using the decomposed \(S_n\) and \(\neg S_n\).
In \textbf{step 2}, 
the \searcherLogo Search Router uses the \(C_\text{current}\) and \(P_n\) to search for \(C_\text{complement}\).
The \resolverLogo Resolver then resolves \(C_\text{current}\) with \(C_\text{complement}\) to produce \(C_\text{resolved}\).
The reasoning complete if: (1) the \(C_\text{resolved}\) determines whether a contradiction exists; 
(2) reach the maximum number of iterations \(I_\text{max}\).
In \textbf{step 3}, \texttt{Aristotle} then concludes the Proof \(D_{S_n}\) and \(D_{\neg S_n}\) based on the Proof Determination.
Using these proofs, \texttt{Aristotle} determines the final answer based on Eq. (\ref{eq:logical_classification}).
Note that two distinct reasoning paths will be implemented: a solid box representing the path starting from \(\neg S_n\), and a dotted box representing the path starting from \(S_n\). 
The complete reasoning process for both two paths, including all iterations are shown in the right part ``\emph{Reasoning Trajectories}''.
}
\label{framework}
\vspace{-4mm}
\end{figure*}

Compared to other forms of general reasoning, logical reasoning \cite{logic-reasoning} stands out as one of the most challenging tasks, as it demands the strictest evidence, arguments, and logical rigor to arrive at sound conclusions or judgments.
Logical reasoning more closely mirrors human-level cognitive processes, making it crucial in high-stakes domains such as mathematical proof generation, legal analysis, and scientific discovery \cite{logic-reasoning1, logic-reasoning2}.
In recent years, numerous studies have investigated how to integrate LLMs into logical reasoning. 
For example, some methods \cite{logic-lm, PAL} use LLMs to translate textual problems into symbolic expressions, which are then addressed by external logic solvers. 
Subsequent work, such as SymbCoT \cite{symbcot}, suggests that LLMs themselves can handle both symbolic translation and logic resolution, thus avoiding potential information loss caused when using external solvers.
While SymbCoT has achieved state-of-the-art (SoTA) performance, the inherent simplicity of CoT's linear reasoning process leaves considerable room for further improvement in LLM-based logical reasoning.

In response, certain research \cite{ToT, got, CR} has applied sophisticated general-purpose reasoning methods (e.g., ToT, GoT) directly to logical reasoning tasks.
Unfortunately, these approaches \emph{\color{blue} largely overlook the inherent structure of logical tasks and fail to effectively integrate logical rules} into the decompose-search-resolve framework, leaving key issues unresolved in both reasoning efficacy and efficiency:

\noindent$\blacktriangleright$
\textbf{From an efficacy perspective}, first, when LLMs decompose logical problems, they often rely on the linguistic token relations rather than the underlying logical structure, leading to disconnected sub-problems and faulty reasoning. 
Specifically, when reasoning hinges on specific logical relationships, neglecting them can result in disjointed sub-problems, breaking the logical chain and ultimately leading to incorrect conclusions.
Furthermore, during the search stage, current path search methods rely heavily on evaluators that may be unreliable, selecting nodes based on possibly flawed logic, causing error propagation through subsequent reasoning steps \cite{evaluator_1, evaluator2}.
For the resolving step, these methods guide LLMs to solve sub-questions with simple text prompts, which frequently contain logical errors, resulting in numerous faulty nodes in the search space \cite{symbcot}. 
These errors propagate through subsequent reasoning steps, causing entire paths to fail and leading to reasoning failure. 
Our preliminary experiments reveal that directly applying SoTA general-purpose reasoning methods with a search mechanism to logical tasks results in significant errors, with 28.4\% for reasoning and 15.0\% for search, as shown in Fig. \ref{intro}.

\noindent$\blacktriangleright$
\textbf{From an efficiency perspective}, these approaches also lead to significant shortcomings.
For example, generating large numbers of incorrect nodes wastes computational resources \cite{sot}. 
Moreover, relying on unreliable evaluators introduces bias into the search process, leading to unnecessary node and path explorations, ultimately reducing efficiency \cite{evaluator_1}.
We note that inefficient logical reasoning systems can significantly undermine their value in practical application scenarios.

To address these challenges, we propose a novel reasoning framework, \textbf{\texttt{Aristotle}}, which effectively tackles the performance and the efficiency bottlenecks in existing logical reasoning tasks by completely integrating symbolic expressions and rules into each stage of the decomposition, search, and resolution.
Fig. \ref{framework} illustrates the overall framework.
Specifically, we first introduce a \textbf{Logical Decomposer} that breaks down the original problem into smaller and simpler components based on its logical structure, reducing the complexity of logical tasks.
We then devise a \textbf{Logical Search Router}, which leverages proof by contradiction to directly search for logical inconsistencies, thereby reducing search errors from unreliable evaluators and minimizing the number of steps required by existing methods.
Finally, we develop a \textbf{Logical Resolver}, which rigorously resolves logical contradictions at each reasoning step, guided by the Logical Search Router.
Overall, \texttt{Aristotle} thoroughly considers the inherent logical structure of tasks, fully incorporating logical symbols into the entire decompose-search-resolve framework. 
This ensures a more logically coherent reasoning process, leading to more reliable final results.

We conducted experiments across multiple logical reasoning benchmarks, where our method surpasses the current SoTA baselines by 4.5\% with GPT-4 and 5.4\% with GPT-4o.
Further analysis revealed that the decomposer, search router, and resolver modules each contributed to:
\textbf{(i)}~reducing task complexity during problem decomposition, leading to improved accuracy in subsequent search and reasoning phases;
\textbf{(ii)}~focusing search efforts on the most direct and relevant paths, which reduced errors and enhanced efficiency;
\textbf{(iii)}~achieving near-perfect logical reasoning accuracy.
Moreover, we observe that \texttt{Aristotle} delivers even greater performance improvements in complex scenarios, such as those with more intricate logical structures or longer reasoning chains.
Overall, this work marks the first successful complete integration of symbolic logic expressions into every stage of an LLM-based reasoning framework (decomposition, search, and resolution), demonstrating that LLMs can perform complete logical reasoning over symbolic structures.

\section{\texttt{\aristotleLogo Aristotle} Architecture}

\vspace{-2mm}

We first formally define the logical reasoning task.
Given a set of premises \(P = \{p_1, p_2, \ldots, p_n\}\), where each \(p_i\) represents a logical statement, a reasoner should derive an answer \(A\) regarding a given statement \(S\). 
The possible answer is true (\(T\)), false (\(F\)), unknown (\(U\)), or self-contradictory (\(SD\)).\footnote{Self-contradictory means a statement can be proved true and false simultaneously.}
The formal definition of each answer can be found in Eq. (\ref{eq:logical_classification}).

As illustrated in Fig. \ref{framework}, \texttt{Aristotle} has an architecture with four modules: \textbf{Translator}, \textbf{Decomposer}, \textbf{Search Router}, and \textbf{Resolver}.

\vspace{-1mm}
\paragraph{\translationLogo Translator.} We use the LLM itself to parse the given premises \(P\) and question statement \(S\) into a symbolic format, which aims to eliminate ambiguity and ensure precision in the logical statement.
We specifically use Logic Programming (LP) language, adopting Prolog's grammar \cite{prolog} to represent the problem as facts, rules, and queries. 
Facts and rules \cite{logic-terminology} are derived from \(P\), while queries are formulated based on the \(S\).
We denote the translated premises (facts and rules) as \(P_t\), and queries as \(S_t\).
The details of the grammar can be found at \ref{logic_grammar}

\vspace{-1mm}
\paragraph{\decomposerLogo Decomposer.} By breaking down the logical statement into a standardized logical form, we can simplify the reasoning process, making it easier to apply formal rules and perform efficient logical calculations.
To achieve this, we use an LLM to transform the parsed premises \(P_t\), and queries \(S_t\) into a standardized logical form through Normalization \cite{cnf} and Skolemization \cite{skolemization}, converting them into Conjunctive Normal Form (CNF)
and eliminates quantifiers, denoted as \(P_n\) and \(S_n\).
For example, the logical rule \(\forall x \, (P(x) \rightarrow Q(x))\) will be decomposed into \(\neg P(x) \lor Q(x)\).

\vspace{-1mm}
\paragraph{\searcherLogo Search Router.}
We adopt the proof-by-contradiction \cite{proof-by-contradiction} approach because it allows us to straightforwardly search for complementary clauses. 
This method reduces search errors and directly targets logical conflicts, making the reasoning process faster and more efficient.
We design a rule-based module to search for the 
clauses \(C_{\text{complement}} \in P_n\) such that \(C_{\text{current}}\) and \(C_{\text{complement}}\) contain complementary terms.
Terms are complementary when they share the same predicate and argument, but have opposite polarity.
For example, if the \(C_{\text{current}}\) is \( \text{P}(\text{x}, \text{True}) \), clauses in the premises that contains \( \text{P}(\text{x}, \text{False}) \) will be found by the Search Router as \(C_{\text{complement}}\), since they are complementary (same predicate \(P\) and argument \(x\) but opposite polarity (True vs. False)).
We will explain how we define the \(C_\text{current}\) in Section \ref{sec:reason_process}.
We include more details about the search strategy in Appendix \ref{search_3_situations} and \ref{alg:full_algo}.

\vspace{-1mm}
\paragraph{\resolverLogo Resolver.} 
To conduct effective step-wise reasoning during proof by contradiction, we adhere to the resolution principle \cite{resolution_principle} as it provides clear and concise instructions to resolve logical conflicts, minimizing the likelihood of reasoning errors.
Specifically, it works by canceling out the complementary terms identified by the Search Router and connecting the remaining terms, a process that will be implemented using an LLM.
Specifically, given two clauses \( C_\text{current} \) and \( C_\text{complement} \), where:
\setlength\abovedisplayskip{3pt}
\setlength\belowdisplayskip{3pt}
\[
C_{\text{current}} = P(x, \text{True}) \lor A
\]
\[
C_{\text{complement}} = P(x, \text{False}) \lor B
\]
Here, \( P(x, \text{True}) \) and \( P(x, \text{False}) \) are complementary terms. The Resolver cancels out them and connects the remaining terms. The resolved clause becomes:
\[
C_{\text{resolved}} =  \text{Resolve}(C_{\text{current}}, C_{\text{complement}}) = A \lor B
\]
If the remaining clause is empty or contradiction (\(\bot\))\footnote{E.g. Resolve \(C_{\text{current}} = (A) \) and \(C_{\text{complement}} = (\neg A)
\) will get an empty clause \(
C_{\text{resolved}} = \bot
\). An empty clause is equivalent to a contradiction.}
, we can conclude the proof and determine the answer, which will be explained in detail in Section \ref{sec:reason_process} at Step 2.

\vspace{-1mm}
\section{Logic-Complete Reasoning Processing}
\label{sec:reason_process}

\vspace{-2mm}
With \texttt{Aristotle}, we now demonstrate how each module comes into play to form the integrated dual-path reasoning process.

\vspace{-2mm}
\paragraph{Step 1: Search Initialization.}

As shown in the step 1 of Fig. \ref{framework}, given the original premises \(P\) and the question statement \(S\), we first translate them into symbolic format \(P_t\) and \(S_t\), and then decompose them into \(P_n\) and \(S_n\), respectively. 

\vspace{-2mm}

\begin{tcolorbox}[boxsep=0pt, left=9pt, right=10pt, top=2pt, bottom=2pt, arc=0pt, fontupper=\linespread{0.7}\selectfont,]
{\footnotesize
\textbf{Translate and Decompose}  \vspace{2pt} \\
$\blacktriangleright$ \textbf{Input:}
\(P\), \(S\)

$\blacktriangleright$ \textbf{Output:}
\(P_n\), \(S_n\), \(C_\text{current}\) = \{\(S_n\), \(\neg S_n\)\}
}
\end{tcolorbox}

\vspace{-2mm}

To implement proof by contradiction, we initialize the current clause \(C_{\text{current}}\) with both \(S_n\) and its negation \(\neg S_n\), denoted as \(
C_{\text{current}} = S_n  
\) and
\(
C_{\text{current}} = \neg S_n  
\).
Considering both \(S_n\) and \(\neg S_n\) is necessary because we need both proofs to scrupulously conclude an answer, which is marked in Eq. (\ref{eq:logical_classification}) and will be explained in detail later in Step 3.

\vspace{-2mm}
\paragraph{Step 2: Search and Resolve.}
\label{search_resolve}

At this stage, two reasoning paths are initiated: one from \(C_\text{current} = S_n\) and the other from \(C_\text{current} = \neg S_n\), initialized in Step 1.
We aim to reach a final answer using proof by contradiction for both paths, iteratively search for complementary clauses and resolve conflicts.
This helps us systematically reach an accurate final answer more quickly.
Specifically for each reasoning path, presented in the Step 2 of Fig. \ref{framework}, the Search Router selects clauses \(C_{\text{complement}} \in P_n\) that are complementary to \(C_\text{current}\).

\vspace{-2mm}

\begin{tcolorbox}[fontupper=\linespread{0.7}\selectfont, boxsep=0pt, left=9pt, right=10pt, top=2pt, bottom=2pt, arc=0pt]
{\footnotesize
\textbf{Search} \vspace{5pt} \\
$\blacktriangleright$ \textbf{Input:}
\(P_n\), \(C_\text{current}\)

$\blacktriangleright$ \textbf{Output:}
\(C_\text{complement}\)
}
\label{statement_determ}
\end{tcolorbox}

\vspace{-2mm}

The Resolver module then applies the resolution rule Resolve(\(C_{\text{current}}\), \(C_{\text{complement}}\)) to produce a new clause \(C_{\text{resolved}}\). 

\vspace{-2mm}

\begin{tcolorbox}[fontupper=\linespread{0.7}\selectfont, boxsep=0pt, left=9pt, right=10pt, top=2pt, bottom=2pt, arc=0pt]
{\footnotesize
\textbf{Resolve} \vspace{5pt} \\
$\blacktriangleright$ \textbf{Input:}
\(C_\text{current}\), \(C_\text{complement}\)

$\blacktriangleright$ \textbf{Output:}
\(C_\text{resolved}\)
}
\label{statement_determ}
\end{tcolorbox}

\vspace{-2mm}

If the \(C_\text{resolved}\) indicates a contradiction or confirms the absence of a contradiction, we then terminate the reasoning process.
If not, we then update \(C_\text{current}\) = \(C_\text{resolved}\) and repeat the Search and Resolve process.
If the process reaches the maximum number of iterations \(I_\text{max}\) and still does not find a contradiction, we conclude that there is no contradiction and terminate the reasoning process.
Given the determination of whether contradiction exists, we then use the formula presented below to formally establish the proof \(D_{S_n}\) (started from \(C_\text{current}\) = \(S_n\)) and \(D_{\neg S_n}\) (started from \(C_\text{current}\) = \(\neg S_n\)) to determine whether \(P_n\) entails either \(S_n\) or \(\neg S_n\).

\vspace{-2mm}

\begin{tcolorbox}[fontupper=\linespread{0.7}\selectfont, boxsep=0pt, left=9pt, right=10pt, top=2pt, bottom=2pt, arc=0pt]
{\footnotesize
\textbf{Proof Determination} \vspace{5pt} \\
\(D_{S_n} = 
\begin{cases} 
P_n \vdash \neg S_n & \text{(\(C_\text{resolved}\) = Contradiction)} \\
P_n \not\vdash \neg S_n & \text{(\(C_\text{resolved}\) = No Contradiction)}
\end{cases}\)
\\
\(D_{\neg S_n} = 
\begin{cases} 
P_n \vdash S_n & \text{(\(C_\text{resolved}\) = Contradiction)} \\
P_n \not\vdash S_n & \text{(\(C_\text{resolved}\) = No Contradiction)}
\end{cases}\)
}
\label{statement_determ}
\end{tcolorbox}

\vspace{-2mm}

\vspace{-2mm}
\paragraph{Step 3: Conclude Answer.}
\label{answer_conclude}
This proof \(D_{S_n}\) and \(D_{\neg S_n}\) can then be used to conclude the truth value \(A\) of \(S\) based on Eq. (\ref{eq:logical_classification}). 
For example, consider a statement \(S\). 
If we get \(D_{S_n}\) = \(P \vdash \neg S\) and \(D_{\neg S_n}\) = \(P \not\vdash S\), 
the combination of \(P \vdash \neg S\) and \(P \not\vdash S\) leads to the conclusion \(A\) that \(S\) is false according to Eq. (\ref{eq:logical_classification}).

\vspace{-2mm}

\begin{tcolorbox}[fontupper=\linespread{0.7}\selectfont, boxsep=0pt, left=9pt, right=10pt, top=2pt, bottom=2pt, arc=0pt]
{\footnotesize
\textbf{Final Answer} \vspace{5pt} \\
$\blacktriangleright$ \textbf{Input:}  \\
\[
D_{S_n} \in \left\{ P_n \vdash \neg S_n, P_n \not\vdash \neg S_n \right\}
\]
\[
D_{\neg S_n} \in \left\{ P_n \vdash S_n, P_n \not\vdash S_n \right\}
\]
$\blacktriangleright$ \textbf{Output:} \\
\[A \in \{\text{True, False, Unknown, Self-Contradictory}\}\]
}
\end{tcolorbox}

\vspace{-2mm}

\vspace{-3mm}
\begin{equation}\small
A = \left\{
\begin{array}{ll}
\text{True,} & P_n \vdash S_n \land P_n \not\vdash \neg S_n \\
\text{False,} & P_n \not\vdash S_n \land P_n \vdash \neg S_n \\
\text{Unknown,} & P_n \not\vdash S_n \land P_n \not\vdash \neg S_n \\
\text{Self-Contradictory,} & P_n \vdash S_n \land P_n \vdash \neg S_n \\
\end{array}
\right.
\label{eq:logical_classification}
\end{equation}

The full algorithm and an example case can be found in Appendix \ref{alg:full_algo} and \ref{example_search_resolve}, respectively.

\begin{table*}[!t]
  \centering
  \fontsize{9}{11}\selectfont 
  \setlength{\tabcolsep}{2mm}
  \begin{tabular}{c l cccc cccc} 
  \toprule
  \multirow{2}{*}{} & \multirow{2}{*}{\textbf{Method}} & \multicolumn{4}{c}{\textbf{GPT-4}} & \multicolumn{4}{c}{\textbf{GPT-4o}} \\
  \cmidrule(r){3-6} \cmidrule(l){7-10} 
   & & ProntoQA & ProofWriter & LogicNLI & Avg & ProntoQA & ProofWriter & LogicNLI & Avg \\
  \midrule
  \multirow{2}{*}{\rotatebox[origin=c]{90}{\textbf{LR}}} & Naive & 77.4 & 53.1 & 49.0 & 59.8 & 89.6 & 48.7 & 53.0 & 63.8 \\
  & CoT & \underline{98.9} & 68.1 & 51.0 & 72.6 & 98.0 & 77.2 & 61.0 & 78.7 \\
  \hline
  \multirow{4}{*}{\rotatebox[origin=c]{90}{\textbf{AR}}} 
  & CoT-SC & 93.4 & 69.3 & 57.3 & 73.3 & \textbf{99.6} & 78.3 & \underline{64.3} & 80.7 \\
  & CR & 98.2 & 71.7 & \underline{62.0} & 77.3 & \textbf{99.6} & 82.2 & 61.0 & \underline{80.9} \\
  & DetermLR & 98.6 & 79.2 & 57.0 & 78.3 & 93.4 & 69.8 & 58.0 & 75.7 \\
  & ToT & 97.6 & 70.3 & 52.7 & 73.5 & 98.6 & 69.0 & 56.7 & 74.8 \\
  \hline
  \multirow{3}{*}{\rotatebox[origin=c]{90}{\textbf{SR}}} & SymbCoT & \textbf{99.6} & \underline{82.5} & 59.0 & \underline{80.4} & \underline{99.4} & \underline{82.3} & 58.7 & 80.1 \\
  & Logic-LM & 83.2 & 79.7 & - & - & 83.2 & 72.0 & - & - \\
  \rowcolor{gray!20} & \bf Ours & \textbf{99.6} & \textbf{86.8} & \textbf{68.3} & \textbf{84.9} & \textbf{99.6} & \textbf{88.5} & \textbf{70.7} & \textbf{86.3} \\
  \specialrule{0em}{-1pt}{-1pt} & & \scriptsize{(+0.0)} & \scriptsize{(+4.3)} & \scriptsize{(+6.3)} & \scriptsize{(+4.5)} & \scriptsize{(+0.0)} & \scriptsize{(+6.2)} & \scriptsize{(+6.4)} & \scriptsize{(+5.4)} \\
  \hline
  \end{tabular}
\vspace{-2mm}
  \caption{Performance on GPT-4 and GPT-4o. The second best score is \underline{underlined} and \textbf{bold} one is the best. In the brackets are the corresponding improvements in between.}
\label{main-1}
\vspace{-2mm}
\end{table*}

\vspace{-1mm}
\section{Experiments}

\vspace{-2mm}
We present the experiment settings, baselines and results in this Section.

\begin{table}[!t]
  \centering
  \fontsize{9}{11}\selectfont 
\setlength{\tabcolsep}{1.5mm}
\begin{tabular}{lcccccccc}
\hline
  & ProntoQA & ProofWriter & LogicNLI & Avg \\
\hline
\multicolumn{5}{l}{$\bullet$ \textbf{\emph{Claude-3.5-Sonnet}}} \\
CoT-SC & \underline{98.0} & \underline{78.5} & 54.3 & \underline{77.0} \\
CR & 88.8 & 57.8 & \underline{57.7} & 68.1  \\
ToT & 92.0 & 69.5 & 46.7 & 69.4 \\
\rowcolor{gray!20}  \bf Ours & \textbf{99.0} & \textbf{86.5} & \textbf{61.3} & \textbf{82.3}  \\
\specialrule{0em}{-1pt}{-1pt} & \scriptsize{(+1.0)} & \scriptsize{(+8.0)}  & \scriptsize{(+3.6)} & \scriptsize{(+5.3)} \\
\hline
\multicolumn{5}{l}{$\bullet$ \textbf{\emph{Llama-3.1-405b}}} \\
CoT-SC & 84.0 & \underline{69.5} & \underline{60.3} & 71.3 \\
CR & \underline{96.0} & 56.3 & 50.7 & 67.7  \\
ToT & \textbf{98.4} & 65.5 & 56.7 & \underline{73.5} \\
\rowcolor{gray!20}  \bf Ours & \textbf{98.4} & \textbf{89.5} & \textbf{69.0} & \textbf{85.6}  \\
\specialrule{0em}{-1pt}{-1pt} & \scriptsize{(+0.0)} & \scriptsize{(+20.0)}  & \scriptsize{(+8.7)} & \scriptsize{(+12.1)} \\
\hline
\end{tabular}%
\vspace{-2mm}
\caption{
Performance by using Claude-3.5-Sonnet and Llama-3.1-405B LLMs.
}
\label{tab:main-1}
\vspace{-2mm}
\label{other-model-2}
\end{table}

\vspace{-2mm}
\subsection{Settings}

\paragraph{LLMs.} We assess the baselines and our method using \textbf{GPT-4} and \textbf{GPT-4o}. We also include \textbf{Claude} and \textbf{LLaMA} to verify whether our method can generalize to different LLMs other than GPT series.

\vspace{-2mm}
\paragraph{Dataset.} We evaluated both the baselines and our method on three carefully selected logical reasoning datasets: \textbf{ProntoQA} \cite{prontoqa}, \textbf{ProofWriter} \cite{proofwriter} and \textbf{LogicNLI} \cite{logicnli}.
These datasets were chosen to reflect increasing levels of difficulty, with ProntoQA being the easiest, ProofWriter moderately complex, and LogicNLI the most challenging due to their intricate logical structures. 
ProntoQA focuses on basic deductive logical relationships, ProofWriter introduces more complex structures such as ``and/or,'' and LogicNLI presents the most intricate reasoning with constructs such as ``either/or'' and ``if and only if''.
This progression enables us to comprehensively evaluate the effectiveness of our method across varying levels of complexity in logical structure.
The details of each dataset can be found in appendix \ref{dataset_details}.

\vspace{-2mm}
\paragraph{Baselines.} 
We compare with a wide range of established baselines. Those baselines can be classified into three main categories.
(1) \textbf{Linear Reasoning} (LR) refers to approaches where the model arrives at an answer through a single-step process, using a straightforward response based on the initial prompt including: 
\emph{Naive Prompting} and 
\emph{CoT} \cite{cot};
(2) \textbf{Aggregative Reasoning} (AR) refers to methods where the model performs reasoning multiple times or aggregates the results to reach a final answer. This includes: 
\emph{CoT-SC} \cite{CoT-SC};
\emph{Cumulative Reasoning} \citep[CR; ][]{CR};
\emph{DetermLR} \cite{DetermLR};
\emph{ToT} \cite{ToT};
(3) \textbf{Symbolic Reasoning} (SR), which engages symbolic expressions and rules in the reasoning framework including: 
\emph{SymbCoT} \cite{symbcot} and
\emph{Logic-LM} \cite{logic-lm}.
More details can be found in Appendix \ref{baseline_details}.

\vspace{-2mm}
\subsection{Main Result}

\vspace{-2mm}
The main results are presented in Table \ref{main-1}, from which we can learn the following observations:

\vspace{-2mm}
\paragraph{Our method consistently outperforms all baselines across the three datasets.}
Specifically, we achieve average improvements over CoT-SC, ToT, CR, and SymbCoT of 11.6\%, 11.4\%, 7.6\%, and 4.5\% on GPT-4, and 5.6\%, 11.5\%, 5.4\%, and 6.2\% on GPT-4o, respectively. These results demonstrate the general advantage of our method over the existing baselines across different datasets.

\vspace{-2mm}
\paragraph{Our method performs even more effectively in complex logical scenarios.} 
We notice in Table \ref{main-1} that our approach does not yield an improvement on the ProntoQA dataset. 
This can be attributed to the relative simplicity of the dataset, where most baselines already achieve high accuracy, leaving limited room for further enhancement. 
However, our improvements are more pronounced on the challenging datasets. Specifically, we achieve a 4.3\% and 6.2\% improvement over the second-best baseline on ProofWriter with GPT-4 and GPT-4o, respectively. 
On the most challenging dataset, LogicNLI, we observe even greater improvements of 6.3\% for GPT-4 and 6.4\% for GPT-4o.
These results highlight the advantages of our method in scenarios involving complex logical structures and increased difficulty.

\vspace{-2mm}
\paragraph{Our method is generalizable across different models.}
In Table \ref{other-model-2}, we present the results for two models (Claude and Llama) outside the GPT series. 
We compare our method with strong baselines that aggregate multiple reasoning paths.
Our method demonstrates similar improvements over the selected strong baseline, highlighting its generalizability across different models.

\vspace{-1mm}
\begin{figure}[!t]
\centering
\resizebox{0.98\columnwidth}{!}{
  \includegraphics[trim=0 0 0 27, clip]{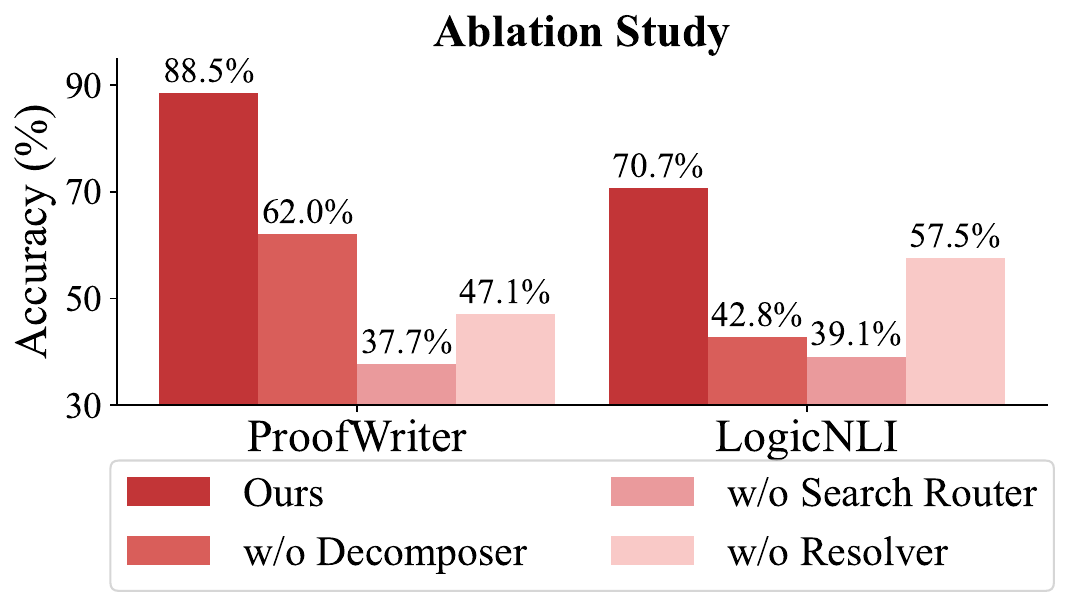}
}
\vspace{-2mm}
\caption{
Ablation results (w/ GPT-4o).
}
\label{ablation}
\vspace{-4mm}
\end{figure}

\vspace{-1mm}
\subsection{Model Ablation}

\vspace{-1mm}
To evaluate the contribution of each module in our framework, we conducted an ablation study by replacing each module individually with simpler alternatives. Specifically, we substitute 
(1) the Decomposer by prompting the LLM for simple decomposition,
(2) the Resolver by prompting the LLM to infer using the given premises, 
and (3) the Search Router by prompting the LLM to search for relevant premises.

The results, shown in Fig. \ref{ablation}, demonstrate that removing any module leads to a significant performance drop, highlighting the importance of each component. 
Notably, replacing the Search Router results in the largest performance decline (50.8\% and 31.6\% for ProofWriter and LogicNLI, respectively), emphasizing the benefits of searching complementary premises under the proof-by-contradiction strategy. 
Besides, the Decomposer has a greater impact than the Resolver on LogicNLI, whereas in ProofWriter, the Resolver plays a more significant role than the Decomposer.
This is because LogicNLI includes more complex logical structures, such as ``either...or...'', ``vice versa'', and ``if and only if'', while ProofWriter primarily involves simpler conjunctions such as ``and'', ``or''. 
As a result, LogicNLI relies more heavily on the Decomposer to break down complex logical statements into simpler forms for optimal performance.

\vspace{-2mm}
\section{Analysis and Discussion}

\vspace{-2mm}
We now take one step further, delving into the underlying working mechanisms of our system.

\vspace{-1mm}
\begin{figure}[!t]
\centering
\resizebox{0.95\columnwidth}{!}{
  \includegraphics{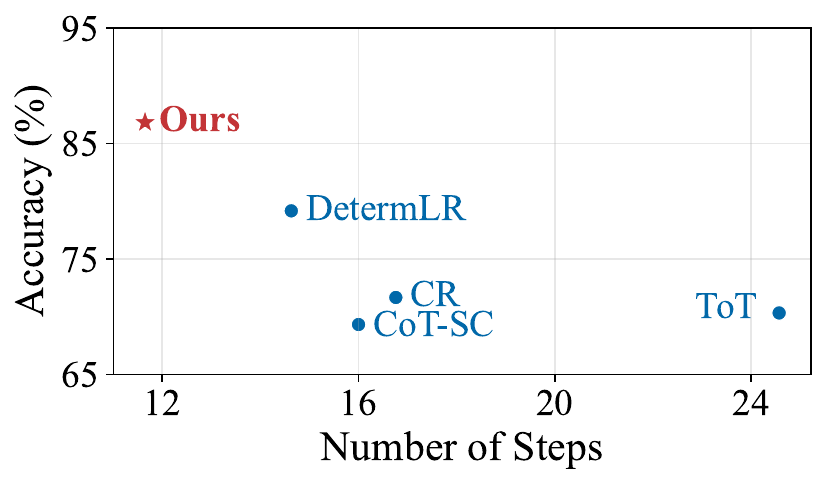}
}
\vspace{-2mm}
\caption{
Accuracy vs. Efficiency on ProofWriter using GPT-4. 
Efficiency is measured as the average number of visited nodes/steps required to solve the problem. The upper-left corner is the optimal point, representing the best performance with the fewest visited nodes.
}
\label{Efficiency}
\vspace{-4mm}
\end{figure}

\vspace{-1mm}
\subsection{Accuracy vs. Efficiency}

\vspace{-1mm}
\paragraph{Our method achieves better reasoning accuracy with higher efficiency.}
Here, we measure the average number of steps or nodes for solving problems in the ProofWriter dataset.
As shown in Fig. \ref{Efficiency}, our method not only achieves the highest accuracy across all baselines but does so with the least number of visited nodes indicating both superior efficacy and efficiency.
Specifically, our method achieves the highest accuracy among all baselines, while visiting only 11.65 nodes on average, reducing the number of nodes visited by 52.6\%, 30.5\%, and 20.4\% compared to ToT, CR, and DetermLR, respectively.
This demonstrates that our approach effectively balances accuracy and computational efficiency.
By directly targeting contradictions, it significantly streamlines the reasoning process, making it both precise and efficient.

\vspace{-2mm}
\subsection{Step-wise Reasoning Accuracy}

\begin{figure}[!t]
\centering
\resizebox{0.98\columnwidth}{!}{
  \includegraphics{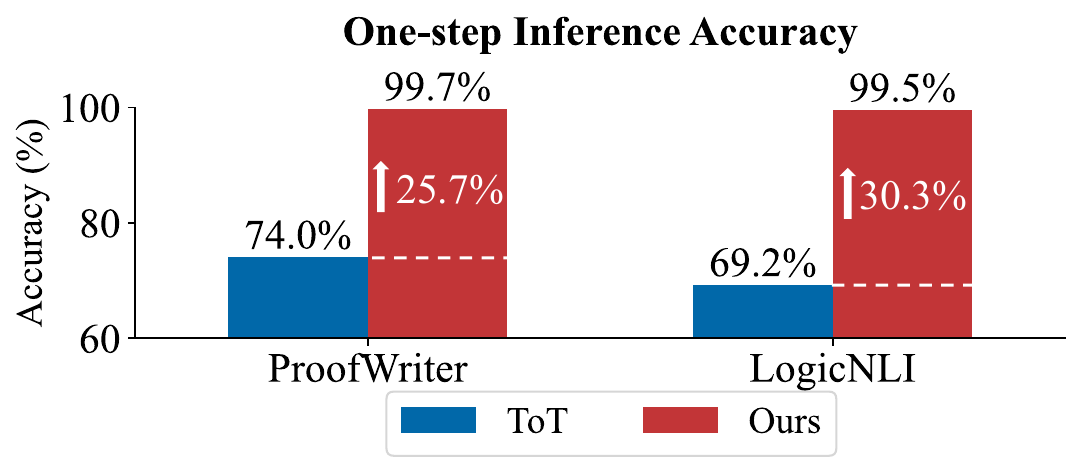}
}
\vspace{-2mm}
\caption{
One-step reasoning accuracy using GPT-4o.
}
\vspace{-4mm}
\label{infer_acc}
\end{figure}

\vspace{-1mm}
\paragraph{The \resolverLogo Resolver can achieve near-perfect accuracy in one-step logical inference.} To understand why our framework is effective, we must also examine its one-step logical reasoning accuracy. 
Since the final answer is derived from these individual inferences (i.e., nodes in ToT and steps in Ours), their accuracy directly impacts the overall performance. 
We compare the one-step reasoning accuracy of our method with that of ToT shown in Fig. \ref{infer_acc}.\footnote{We randomly sample 100 cases with manual evaluation.}
ToT demonstrated around 70\% accuracy, which is consistent with prior research showing that LLMs can sometimes introduce logical errors. 
In contrast, our \texttt{Aristotle} achieved near-perfect accuracy in one-step inference, underscoring the effectiveness of the Resolver module’s use of the resolution principle.
This is because the resolution principle provides a systematic and logically rigorous way to resolve contradictions, simplifying the reasoning process compared to methods that rely on LLMs to reason from previous steps and multiple premises.

\vspace{-2mm}
\subsection{Search Error}
\begin{figure}[!t]
\centering
\resizebox{0.98\columnwidth}{!}{
  \includegraphics{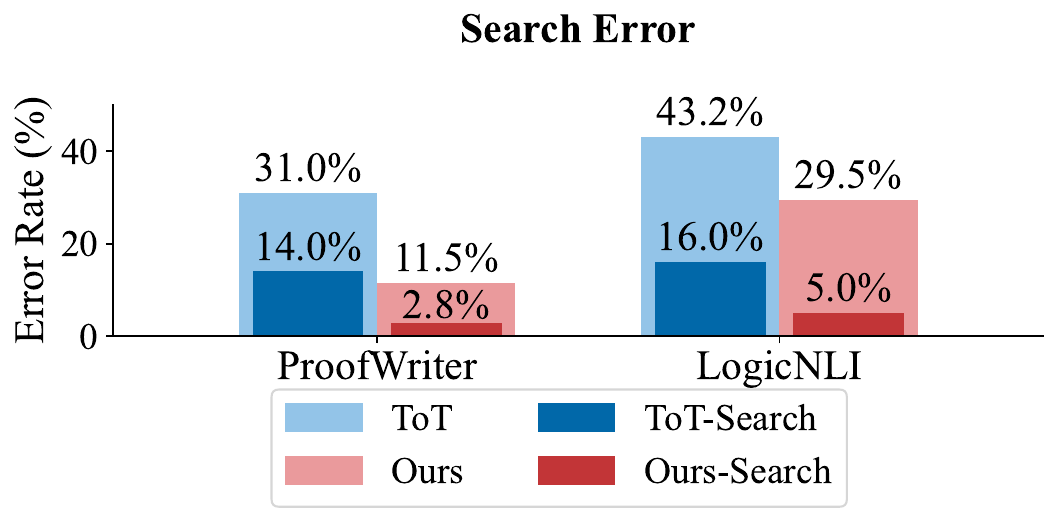}
}
\vspace{-2mm}
\caption{
The outer bar shows the overall error rate. The inner bar represents the proportion of the error caused by the search strategy.
}
\vspace{-2mm}
\label{search_error}
\end{figure}

\vspace{-1mm}
\paragraph{Our method effectively reduces errors from the search strategy.} Apart from one-step logical inference, the search router also plays a crucial role. Previous research has shown that methods involving an evaluator to guide the search tend to underperform as the evaluator can be unreliable and may mislead the reasoning process, resulting in incorrect answers. 
We assess the search error\footnote{In ToT, search errors occur when the evaluator selects a logically flawed node for expansion. In contrast, in our method, search errors arise when either the complementary clause is not found or a non-complementary clause is selected. The evaluation process is conducted manually.}, as shown in Fig. \ref{search_error}. 
Our search strategy significantly reduces errors, lowering them by 11.2\% in ProofWriter and 9.0\% in LogicNLI. 
This demonstrates that our logic-based search approach outperforms LLM self-evaluation, effectively addressing the limitations posed by unreliable evaluators in logical reasoning.
An explanation is that our method simplifies the search process by focusing on identifying complementary clauses, a task with clear definitions and rules that an LLM can easily follow with a few examples. 
In contrast, having an LLM evaluate logical inferences, such as in ToT, requires complex judgments, making it more prone to errors \cite{evaluator_1, evaluator2}.

\vspace{-1mm}
\subsection{Complex Reasoning}

\begin{figure}[!t]
\centering
\resizebox{0.98\columnwidth}{!}{
  \includegraphics{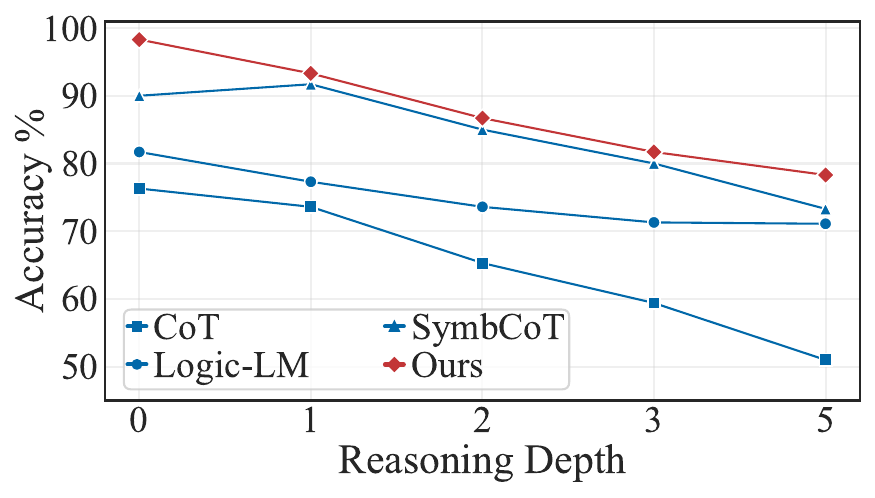}
}
\vspace{-2mm}
\caption{
Studying the effect of reasoning depth with GPT-4 on ProofWriter. 
}
\label{depth}
\vspace{-4mm}
\end{figure}

\vspace{-1mm}
\paragraph{Our method demonstrates a clear advantage in handling problems of increasing difficulty.} 
We evaluate accuracy across different reasoning difficulties in ProofWriter, as shown in Fig. \ref{depth}. Our method consistently outperforms others at all depths, maintaining superior accuracy. It excels particularly at moderate and challenging depths, surpassing baselines like SymbCoT and Logic-LM. Even as other methods struggle at higher depths, our approach remains robust, demonstrating better scalability and resilience to problem difficulty.
This effectiveness is due to two main factors: (1) using the resolution principle minimizes errors at each step and prevents them from compounding, and (2) streamlining the reasoning process reduces steps, lowering the likelihood of error accumulation.

\paragraph{The cost scaling is stable with increased reasoning depth.}
To complement the performance analysis on different reasoning depth, we also address concerns about the scalability of our framework when applied to tasks requiring deeper reasoning involving greater reasoning depth. 
Specifically, we provide a detailed examination of the marginal computational costs and token usage associated with extended reasoning chains. 
Aristotle is designed to handle increased reasoning depth in an efficient and controlled manner.

Each reasoning step in our framework consists of two operations. 
First, a search operation is performed using a deterministic rule-based method, which introduces minimal computational overhead and does not contribute additional token usage as reasoning depth increases. 
Second, a resolve operation is executed by the resolver module, which processes one reasoning step at a time and is the only component contributing to token-based computational cost.

This modular design ensures that total token usage grows \textit{linearly} with the number of reasoning steps, keeping the marginal cost per additional step low and predictable. 
To empirically validate the efficiency and stability of our approach, we analyzed token usage for individual reasoning steps across two benchmark datasets, ProofWriter and LogicNLI, as shown in Table \ref{tab:token_stats}.
The statistics indicate that token usage per reasoning step is highly consistent, exhibiting low standard deviation and coefficient of variation. 
Moreover, the absolute token cost per step remains modest, ensuring that deeper reasoning does not impose a significant computational burden. Given the stable, linear scaling of token consumption and the low per-step cost, our framework maintains its efficiency even for tasks requiring extended reasoning chains. 

\begin{table}[t]
\centering
\resizebox{\columnwidth}{!}{%
{\fontsize{7}{9}\selectfont
\begin{tabular}{lccc}
\toprule
\textbf{Dataset} & \textbf{Avg. TU} & \textbf{Std. Dev.} & \textbf{CV} \\
\midrule
ProofWriter & 3076.8 & 19.1 & 0.71\% \\
LogicNLI    & 2071.1 & 26.4 & 0.85\% \\
\bottomrule
\end{tabular}
}
}
\caption{Token usage statistics per reasoning step. TU denotes Token Usage, Std. Dev. denotes Standard Deviation, and CV denotes Coefficient of Variation.}
\label{tab:token_stats}
\end{table}

\begin{figure}[!t]
\centering
\resizebox{0.98\columnwidth}{!}{
  \includegraphics{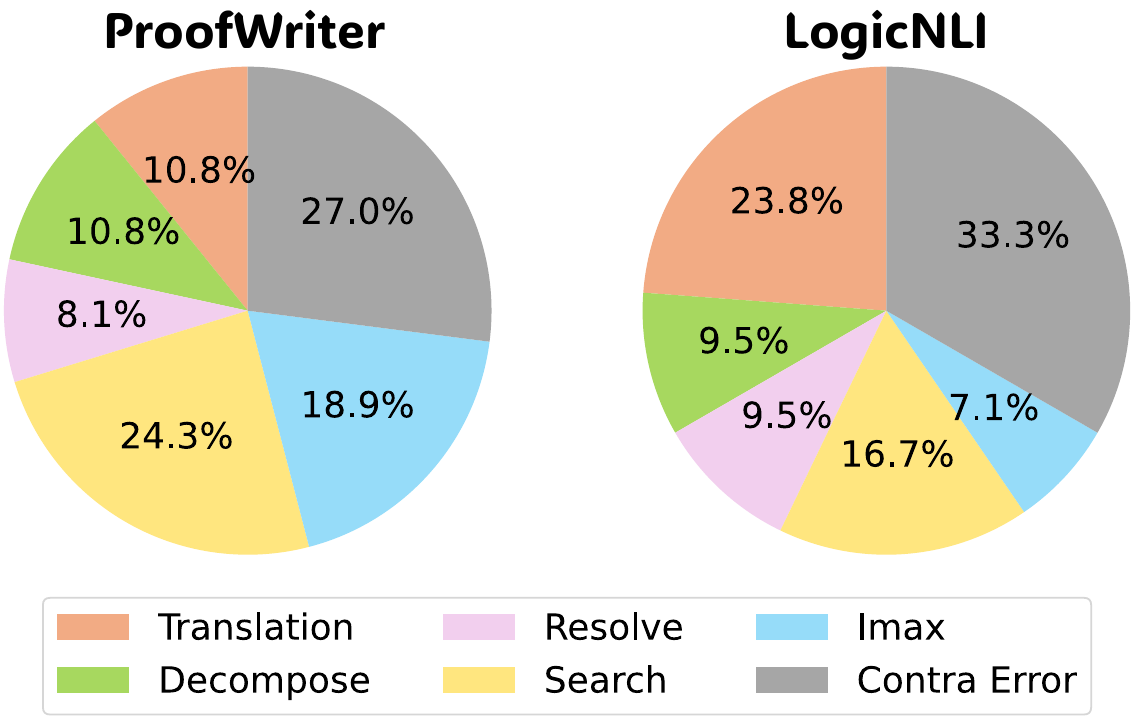}
}
\vspace{-2mm}
\caption{
Error analysis on ProofWriter and LogicNLI with GPT-4o.
}
\label{error_analysis}
\end{figure}
\vspace{-1mm}

\vspace{-1mm}
\subsection{Error Analysis}

\vspace{-1mm}

To thoroughly understand the limitations of our framework, we conduct a manual error analysis on the ProofWriter and LogicNLI datasets using GPT-4o. The detailed error statistics are presented in Fig. \ref{error_stats}.

The majority of errors stem from the \emph{Contradiction Error}, primarily due to flaws in dataset construction. The next most common source is the Search module, where complementary clauses exist but are not retrieved.
This is often due to unexpected symbols (e.g., LaTeX code) in the LLM’s output that disrupt regular expression matching.

\emph{Translation and Decomposition} errors are the third largest category. These occur when the LLM struggles to parse complex logical relationships or convert them into the correct symbolic form. Such errors are more prevalent in LogicNLI, which features more intricate constructs (e.g., "either...or..." and "if and only if") compared to the simpler logic in ProofWriter (e.g., "and," "or").

Another notable source of error is \emph{Insufficient Iterations}, where the reasoning process terminates prematurely, often concluding "No contradiction" when further iterations might reveal one. 
While increasing the iteration threshold could mitigate this, it must be balanced against computational efficiency.

Finally, \emph{Resolution errors}, often due to incorrect variable instantiations, are relatively infrequent. 
This is because logical statements have been reduced to a simple conjunctive normal form (CNF) by this point, making them easier to interpret, and the resolution principle offers clear guidance for resolving inconsistencies.

Potential improvements include incorporating more targeted in-context learning (ICL) examples to enhance translation and resolution. 
Besides, enhancing regular expression patterns to accommodate a wider range of syntactic variations could further reduce search errors. Additionally, tuning iteration limits would help achieve a better balance between accuracy and computational efficiency. More details of error analysis can be found at Appendix \ref{error_stats}.


\vspace{-1mm}
\section{Related Work}

\vspace{-2mm}
Enhancing the logical reasoning of LLMs to achieve human-like levels has been a key focus in recent research \cite{logic-reasoning, human-performance}. 
Existing approaches can be broadly categorized into the following:

\vspace{-2mm}
\paragraph{Linear Reasoning.} These approaches involve prompting LLMs once to emulate human reasoning in a sequential manner. 
The representative is the CoT method \cite{cot}, which guides the model to generate reasoning steps linearly. 
Building upon CoT, SymbCoT \cite{symbcot} incorporates symbolic expressions and rules into the linear reasoning process to achieve more rigorous logical reasoning.
However, these methods lack any searching or backtracking mechanism to avoid flawed reasoning paths, leading to suboptimal performance.
In this paper, we thus consider the searching-based reasoning framework.

\vspace{-2mm}
\paragraph{Sampling Methods.} These methods obtain the final answer through samplings to enhance reasoning diversity and accuracy \cite{CoT-SC, complex-based-sc, selfcheckgpt}. 
This involves running the reasoning process multiple times, with the increased diversity of reasoning paths contributing to better results. 
However, they do not resolve the underlying issue of flawed logical reasoning inherent to the LLM, which is resolved by the resolution principle \cite{resolution_principle} in our framework.

\vspace{-2mm}
\paragraph{Iterative Reasoning with Search.} These methods rely on an evaluator to search for different reasoning paths to avoid flawed nodes. 
Techniques such as ToT \cite{ToT}, and its variances \cite{beam-search, CR, DetermLR}, generate multiple thoughts during the reasoning. 
An evaluator repetitively selects the most probable paths to expand to the next level until the final answer.
However, the evaluator may not be reliable \cite{evaluator_1, evaluator2}, potentially selecting incorrect nodes for expansion and propagating errors.
This paper proposes a search mechanism that relies on matching symbolic logic, avoiding the use of an unreliable evaluator.

\vspace{-2mm}
\paragraph{Reasoning Relying on External Tools.} Here LLMs often involve integrating well-developed rule-based reasoning engines, where LLMs act as mere translators, converting the natural language of reasoning questions into specific symbolic forms to be processed by external rule-based engines. 
Examples include Logic-LM \cite{logic-lm}, LINC \cite{linc} and PAL \cite{PAL}. 
The limitation of this approach lies in the strict formatting requirements of external logic resolver; 
LLMs inevitably introduce syntax errors during translation, leading to failures in the reasoning process.
Fortunately, the success of SymbCoT \cite{symbcot} preliminarily demonstrates that enabling LLMs to perform logical reasoning based on symbols is feasible and promising. 
In our paper, we further prove that symbolic logic expressions can be fully integrated into all processes of the reasoning framework, including decomposition, search, and inference, thereby demonstrating that LLMs themselves can completely achieve high-level symbolic logical reasoning.

\vspace{-2mm}
\section{Conclusion}

\vspace{-2mm}
We presented \texttt{Aristotle}, a logic-complete reasoning framework designed to tackle the challenges of logical reasoning,  
which comprehensively integrates symbolic expressions and logical rules into three core components: Logical Decomposer, Logical Search Router, and Logical Resolver.  
These modules streamline the reasoning process by reducing the task complexity, minimizing the search errors, and rigorously resolving logical contradictions. 
Our experiments show that our method consistently outperforms state-of-the-art frameworks in both accuracy and efficiency significantly.

\vspace{-2mm}
\section*{Acknowledgements}
\vspace{-1mm}
This work is supported by the Ministry of Education, Singapore, under its MOE AcRF TIER 3 Grant (MOE-MOET32022-0001).


\section*{Potential Limitations}
\vspace{-1mm}
Our method faces two potential limitations. 
First, the reasoning process relies on the quality of translation and decomposition. 
However, even with a few-shot approach, LLMs cannot always guarantee that these processes are fully correct.
Future work could consider more advanced methods to guarantee the quality of these processes, such as fine-tuning.
Secondly, our reasoning approach requires that all necessary information is explicitly stated in the premises. 
If any implicit information or assumptions exist, our method may fail to capture them, leading to incorrect reasoning.
A more detailed analysis of this limitation is provided in Appendix \ref{appendix:limitation_analysis}.
Nevertheless, there are existing methods that explore how to make implicit information explicit. 
Future work could integrate those methods into this framework to address this limitation.

\vspace{-2mm}
\section*{Ethics Statement}
\vspace{-1mm}
The datasets employed in our research were carefully selected from publicly available or ethically sourced materials, and we take deliberate steps to identify and reduce biases within these datasets to uphold fairness in our outcomes. 

\vspace{-3mm}

\bibliography{custom}
\bibliographystyle{acl_natbib}

\appendix

\clearpage


\section{Logical Grammar}
\label{logic_grammar}

\textbf{Facts}: A fact is an assertion about specific individuals in the domain. 
It involves a predicate applied to constants (not variables) and states that a particular relationship holds between these individuals. 
For example, \(\text{Sees}(\text{Jack}, \text{Tom}, \text{True})\) asserts Jack sees Tom.

\textbf{Rules}: A rule delineates a logical relationship between predicates and forms an integral component of the domain's terminological knowledge. These rules typically incorporate variables and are universally quantified, ensuring their applicability across all relevant instances within the domain. 
Rules can involve logical connectors such as "and" (\(\land\)), "or" (\(\lor\)), "either...or..." (exclusive or, \(\oplus\)), and "not" (\(\neg\)), appearing on both sides of the implication (\(\rightarrow\)) or biconditional (\(\leftrightarrow\)) operators. 
For instance, the rule 
\setlength\abovedisplayskip{3pt}
\setlength\belowdisplayskip{3pt}
\[
\forall x, y \left( \text{Sees}(x, y) \rightarrow \text{Knows}(x, y) \right)
\]
asserts that for all individuals \( x \) and \( y \), if \( x \) sees \( y \), then \( x \) knows \( y \).

\textbf{Query}: A query is a fact that needs to be proven based on the given facts and rules.

\section{Three potential situations for searching complement}
\label{search_3_situations}
When searching for a complementary clause during the resolution process, three potential situations may arise. 
\begin{compactitem}

    \item[1)] If exactly one complementary clause is found the resolution will be directly implemented. 

    \item[2)] If multiple complementary clauses are identified, we prioritize the shorter clauses, while saving the remaining clauses as backup options. 
    In cases where the current reasoning path cannot find any further complementary clauses before reaching the predefined maximum number of iterations, we will backtrack and attempt to use these backup clauses.

    \item[3)] If no complementary clause is found initially, we will backtrack to the backup list. 
    Should the backup list also be exhausted, we will conclude the reasoning process by determining the result as "No contradiction found." 

\end{compactitem}

This approach ensures a structured and efficient search and resolution process, while also accounting for cases where multiple or no complementary clauses are found, improving overall search robustness.

\section{Conjunctive Normal Form}
\label{cnf}
Conjunctive Normal Form (CNF) is a standardized way of expressing logical statements in Boolean logic. 
In CNF, a formula is composed of a conjunction (AND, denoted as \(\land\)) of clauses, where each clause is a disjunction (OR, denoted as \(\lor\)) of literals. 
A literal is either a variable or its negation. For example, the logical statement \((A \lor \neg B) \land (C \lor D)\) is in CNF. 
Each group within the parentheses is a clause, and the entire expression is the conjunction of these clauses. 
The reason we need logical statements to be in CNF to conduct resolution is that resolution is a fundamental inference rule in automated theorem proving. 
It works by finding pairs of clauses where one contains a literal and the other contains its negation. 
These pairs can then be combined to eliminate the opposing literals, simplifying the overall logical expression. 
Since CNF breaks down complex statements into smaller, manageable components of ANDs and ORs, it allows the resolution rule to systematically and efficiently simplify or refute logical expressions, thus enabling automated reasoning systems to solve problems.

\section{Dataset Specifications}
\label{dataset_details}

\textbf{ProntoQA} is a synthetic dataset designed to assess models' deductive reasoning abilities. For our evaluation, we use the most difficult 5-hop subset. Each question in this dataset requires verifying whether a given claim is "True" or "False" based on the provided premises. The dataset focuses on basic logical relationships. For instance, given "X is Y", "Y is Z", determining whether "X is Z".

\noindent \textbf{ProofWriter} is a popular dataset for logical deductive reasoning. The dataset has five subsets with different reasoning depths (from 1 to 5). We use the hardest depth-5 subset for evaluation. And the context in this dataset contains more challenging logical relationships such as the combination of "and" and "or".

\noindent \textbf{LogicNLI} is a challenging NLI-style dataset that effectively disentangles the target logical reasoning from commonsense inference. In addition to common logical relationships such as "and" and "or", it presents the most difficult relationships among the three datasets, such as "either...or...", "vice versa", and "if and only if".

The test sizes are 500 for ProntoQA, 600 for ProofWriter, and 300 for LogicNLI, respectively.

\section{Evaluation on Real-world Dataset}
\label{appendix:limitation_analysis}

\begin{table}[t]
\centering
{\fontsize{11}{12}\selectfont
\begin{tabular}{lcc}
\toprule
\textbf{Model} & \textbf{FOLIO} & \textbf{LogiQA} \\
\midrule
CoT        & 75.0 & 65.6 \\
Aristotle  & 76.5  & 31.2 \\
\bottomrule
\end{tabular}
}
\caption{Performance on Real-world's Benchmark}
\label{tab:folio_logiqa}
\end{table}

The ability to solve real-world problems is important, as it reflects a model’s applicability beyond synthetic or constrained settings. Numerous benchmarks \cite{folio, multi-logieval, logieval, logicbench, logiqa, reclor, gridpuzzle} have been introduced to evaluate this capability by incorporating tasks that require implicit reasoning, background knowledge, and commonsense understanding.

We include an evaluation on real-world datasets FOLIO \cite{folio} and LogiQA \cite{logiqa}, to provide a deeper understanding of Aristotle. As shown in Table~\ref{tab:folio_logiqa}, Aristotle's performance is suboptimal. This is primarily due to its design: it operates on explicitly stated premises and deliberately avoids relying on unstated assumptions or external background knowledge. However, real-world scenarios frequently depend on such implicit information and commonsense reasoning, making these capabilities essential for robust performance.

While Aristotle promotes clarity and precision in logical inference, it may not fully capture the complexities inherent in real-world tasks—a limitation we have acknowledged in the main paper. Notably, Aristotle’s performance on LogiQA is significantly worse than on FOLIO. Our qualitative analysis reveals that this discrepancy stems from LogiQA’s greater reliance on commonsense knowledge and implicit assumptions, which makes it less compatible with Aristotle’s strictly premise-driven reasoning approach.

To address this, our plan is to incorporate external knowledge to better handle such scenarios in future work. For instance, information retrieval from the internet or commonsense knowledge graphs could supplement the explicit premises with the necessary implicit knowledge for reasoning. This integration would enable our method to leverage background information that is not explicitly provided, improving its applicability and performance in solving real-world tasks.

\section{Error Analysis}
\label{error_stats}



Here, we present a more detailed error analysis.

\subsection{False Contradiction}
\label{false_contra_explain}

False contradiction refers to when the method identifies a contradiction when none should exist, leading to an incorrect final answer. 
This issue often arises in cases where the ground truth of a problem is false. 
For example, when the ground truth is false, we should find a contradiction when reasoning from the negation of the statement and no contradiction when reasoning from the original statement, as outlined in the equation \ref{eq:logical_classification}.

However, our method sometimes finds a contradiction even when reasoning from the original statement, altering the final answer and producing errors. 
This should be due to the way datasets are constructed. 
When the ground truth is false, for instance, the dataset may be built such that the false statement is provable, but the construction process might fail to ensure that the true statement is not provable. 
This oversight results in both the true and false values being provable, making the problem self-contradictory. 
This situation occurs more frequently when the ground truth is either true or false, suggesting that the dataset did not fully account for the exclusive relationship between proving one value and excluding the other.

In an ideal dataset construction where the ground truth is true or false, the premises should only allow for the ground truth to be provable, while any other possible answers should be logically excluded. 
That is, if a statement is provably true, it should be impossible to prove its negation, and vice versa. 
Ensuring this exclusivity is crucial for logical consistency.

That said, it's important to note that these False Contradictions represent only a small portion of the overall dataset.
While this issue can affect some instances, it doesn't significantly undermine the dataset's overall effectiveness for testing reasoning models.

\subsection{Resolver Instantiation Error}
An instantiation error occurs when a resolver incorrectly substitutes a variable, leading to an inaccurate conclusion. For example, given two clauses: \(Smart(Gary, False)\) and \(Smart(Gary, True) \lor Nice(x, False)\), the correct resolution would recognize that \(Smart(Gary, False)\) directly complements \(Smart(Gary, True)\), resulting in the simplified clause \(Nice(x, False)\) without needing to instantiate `x`. 
However, if the resolver mistakenly instantiates "x" as "Gary," the clause changes to \(Nice(Gary, False)\), which is more specific than necessary. 

This error restricts the generality of the conclusion, as the correct clause \(Nice(x, False)\) is intended to apply to any individual, not just "Gary." 
Such improper instantiation can lead to faulty reasoning in subsequent steps, where conclusions might be incorrectly drawn because the reasoning process has been prematurely narrowed to a specific case. 
Ensuring that instantiation only occurs when necessary can help prevent these errors and maintain the validity of logical deductions.

\section{Baselines}
\label{baseline_details}

Here we illustrate the details of each baseline.

\paragraph{Naive Prompting}
Naive Prompting refers to a basic prompting technique where a model is given a question or task without any complex instructions or intermediate steps. The model is expected to output a direct answer based on its existing knowledge. In this approach, the reasoning process is implicit, and the model simply leverages its pre-trained knowledge to respond without additional structured reasoning or step-by-step guidance.

\paragraph{Chain-of-Thought (CoT)}
Chain-of-Thought (CoT) prompting is a more advanced prompting strategy that encourages the model to generate intermediate reasoning steps before arriving at a final answer. Instead of asking the model for an immediate response, the prompt guides the model to break down the reasoning process into smaller, logical steps. This allows the model to engage in more thoughtful problem-solving and often leads to better performance on tasks requiring multi-step reasoning \cite{cot}.

\paragraph{Chain-of-Thought with Self-Consistency (CoT-SC)}
Chain-of-Thought with Self-Consistency (CoT-SC) improves upon the standard CoT method by running the chain-of-thought reasoning process multiple times independently. Instead of producing just one reasoning chain per query, the model generates multiple chains for the same task. After running these different reasoning processes, the final answer is determined by applying majority voting on the outputs. This ensures that the model selects the answer that is most consistent across multiple reasoning attempts, which helps reduce variability and errors caused by randomness or incorrect intermediate steps in any single chain\cite{CoT-SC}.

\paragraph{Cumulative Reasoning (CR)}
Cumulative Reasoning builds on the idea that reasoning can be improved over successive iterations. The model does not simply reach a conclusion in one step, but rather, the reasoning evolves across multiple stages or passes. In this process, intermediate results are used as building blocks for the final solution, allowing the model to accumulate information and refine its reasoning step by step \cite{CR}.

\paragraph{DetermLR}
DetermLR is a reasoning approach that rethinks the process as an evolution from indeterminacy to determinacy. It categorizes premises into determinate (clear) and indeterminate (uncertain) types, guiding models to convert indeterminate data into determinate insights. The approach uses quantitative methods to prioritize relevant premises and employs reasoning memory to store and retrieve historical reasoning paths. This helps streamline the reasoning process, progressively refining the model’s understanding to produce more determinate and accurate conclusions \cite{DetermLR}.

\paragraph{Tree-of-Thought (ToT)}
Tree-of-Thought (ToT) is a framework that uses a tree-like structure for reasoning. Instead of generating a single chain of thought, the model explores multiple reasoning pathways in parallel, branching out into different possible solutions. The tree structure allows the model to evaluate and prune different paths, keeping only the most promising routes to reach the correct solution. This approach is particularly useful for problems where multiple reasoning paths can lead to the answer, allowing for exploration and selection of the best path \cite{ToT}.

\paragraph{SymbCoT}
SymbCoT integrates symbolic expressions and rules into the chain-of-thought (CoT) process. It translates natural language input into symbolic representations, allowing the model to reason based on these symbolic expressions. The LLM then applies symbolic rules to process and analyze the information, enhancing its ability to handle tasks that require formal reasoning and structured problem-solving \cite{symbcot}.

\paragraph{Logic-LM}
Logic-LM translates natural language input into a symbolic format and then applies a rule-based logical engine to perform reasoning. This approach leverages formal logic rules to process and analyze the symbolic representation, enabling more structured and precise reasoning, particularly for tasks that require strict logical inferences \cite{logic-lm}.

\onecolumn
\section{Full Algorithm}
\label{alg:full_algo}
\RestyleAlgo{algcompatible}
\begin{algorithm}[H]
\caption{Methodology}
\SetAlgoLined
\fontsize{9}{11}\selectfont
\KwIn{Premises $P$, Question Statement $S$, LLM $p_\theta$, Translator $T()$, Decomposer $D()$, Search Router $SR()$, Resolver $R()$, Search Round Limit $S$}
$P_t, S_t \leftarrow T(P, S)$ \tcp*{Translate the given premises and statement}
$P_n, S_n \leftarrow D(P_t, S_t)$ \tcp*{Decompose the translated premises and statement}
$C_{\text{current\_list}} \leftarrow [S_n, \neg S_n]$ \tcp*{Initiate search with $S_n$ and its negation}
$Search\_round \leftarrow 0$\;

\ForEach{$C_{\text{current}}$ \emph{in} $C_{\text{current\_list}}$}{
    \While{$Search\_round < S$}{
        $C_{\text{searched\_list}} \leftarrow SR(C_{\text{current}}, P_n)$ \tcp*{Search for complementary clause}
        $num\_searched\_C \leftarrow \text{len}(C_{\text{searched\_list}})$\;

        \eIf{$num\_searched\_C >= 1$}{ 
            $C_{\text{searched}} \leftarrow C_{\text{searched\_list}}[0]$\;
        }{
            $C_{\text{searched}} \leftarrow C_{\text{searched\_list}}.\text{pop}(0)$\;
        }

        $cache \leftarrow \{P_n: C_{\text{searched\_list}}\}$\ \tcp*{If more than one \(C_\text{current}\), save in $cache$}

        \If{$num\_searched\_C == 0$}{ 
            \If{$\text{cache}[C_{\text{current}}]$ is not empty}{
                $C_{\text{current}} \leftarrow \text{next}(\text{iter}(\text{cache}))$\ \tcp*{If no \(C_current\) found, search from $cache$}
                $C_{\text{searched}} \leftarrow \text{cache}[C_{\text{current}}].\text{pop}(0)$\;
            } 
            \Else{ 
                \If{Start from $S_n$}{
                    $D_{S_n} \leftarrow P \not\vdash \neg S $\ \tcp*{If $cache$ is empty, make conclusion}
                }
                \If{Start from $\neg S_n$}{
                    $D_{\neg S_n} \leftarrow P \not\vdash S$\;
                }
                \textbf{break}\;
            }
        }

        $C_{\text{resolved}} \leftarrow R(C_{\text{current}}, C_{\text{searched}})$\;
        \If{$C_{\text{resolved}} == \text{`Contradiction'}$}{
            \If{Start from $S_n$}{
                $D_{S_n} \leftarrow P \vdash \neg S $\ \tcp*{If contradiction is found, make conclusion};
            }
            \If{Start from $\neg S_n$}{
                $D_{\neg S_n} \leftarrow P \vdash S$\;
            }
            \textbf{break}\;
        }
        \Else{
            $P_n \leftarrow P_n \cup \{C_{\text{resolved}}\}$\ \tcp*{Append  \(C_\text{resolved}\) on \(P_n\)}
        }
        $C_{\text{current}} \leftarrow C_{\text{resolved}}$\;
    }
    \If{Start from $S_n$}{
        $D_{S_n} \leftarrow P \not\vdash \neg S $\ \tcp*{If no contradiction found and reach max iterations, make conclusion}
    }
    \If{Start from $\neg S_n$}{
        $D_{\neg S_n} \leftarrow P \not\vdash S$\;
    }
}
\end{algorithm}

\clearpage
\section{Case Study}
\label{example_search_resolve}

Given the premises, we need to determine whether the question statement \(S\) "Dave is not nice" is true/false/unknown/self-contradictory.
We first start the first reasoning path from \(C_\text{current}\) = \(S_n\).
\(C_\text{complement}\) is the complementary clause found by the Search Router from \(P_n\).

\vspace{-2.5mm}

\begin{tcolorbox}[title={Logical Resolution Steps}, colframe=blue!75!black, colback=white, fonttitle=\bfseries, breakable]

\textbf{Translated and Decomposed Premises \(P_n\):}

\begin{enumerate}
\item If someone is green then they are nice ::: \(\forall x \left( \text{Green}(x, \text{False}) \lor \text{Nice}(x, \text{True}) \right)\)
\item If someone is smart then they are green ::: \(\forall x \left( \text{Smart}(x, \text{False}) \lor \text{Green}(x, \text{True}) \right)\)
\item Dave is smart ::: \(\text{Smart}(\text{Dave}, \text{True})\)
\end{enumerate}

Question Statement \(S_n\): Dave is not nice ::: Nice(Dave, False)

\medskip

\(C_{\text{current}}:\quad \text{Nice}(\text{Dave}, \text{False})\)

\bigskip

\textbf{Resolution Steps:}

\begin{enumerate}
\item \textbf{First Resolution:}

\[
\text{Resolve}\left(C_{\text{current}} = \text{Nice}(\text{Dave}, \text{False}),\ C_{\text{complement}} = \forall x \left( \text{Green}(x, \text{False}) \lor \text{Nice}(x, \text{True}) \right)\right)
\]

\begin{itemize}
\item \textit{Instantiate} \(C_{\text{complement}}\) \textit{for} \( x = \text{Dave} \):

  \[
  \text{Green}(\text{Dave}, \text{False}) \lor \text{Nice}(\text{Dave}, \text{True})
  \]

\item \textit{Resolve with} \(C_{\text{current}} = \text{Nice}(\text{Dave}, \text{False})\):

  \[
  \begin{aligned}
  &\text{Nice}(\text{Dave}, \text{False}) \\
  &\quad \text{and} \\
  &\text{Green}(\text{Dave}, \text{False}) \lor \text{Nice}(\text{Dave}, \text{True})
  \end{aligned}
  \]

\item \textit{Since} \(\text{Nice}(\text{Dave}, \text{False})\) \textit{contradicts} \(\text{Nice}(\text{Dave}, \text{True})\), \textit{the new resolved clause is}:

  \[
  C_{\text{resolved}} = \text{Green}(\text{Dave}, \text{False})
  \]

\item \textit{Update \(C_\text{current}\) to \(C_\text{resolved}\)}:

  \[
  C_{\text{current}} = C_{\text{resolved}} = \text{Green}(\text{Dave}, \text{False})
  \]

\end{itemize}

\medskip

\item \textbf{Second Resolution:}

\[
\text{Resolve}\left(C_{\text{current}} = \text{Green}(\text{Dave}, \text{False}),\ C_{\text{complement}} = \forall x \left( \text{Smart}(x, \text{False}) \lor \text{Green}(x, \text{True}) \right)\right)
\]

\begin{itemize}
\item \textit{Instantiate} \(C_{\text{complement}}\) \textit{for} \( x = \text{Dave} \):

  \[
  \text{Smart}(\text{Dave}, \text{False}) \lor \text{Green}(\text{Dave}, \text{True})
  \]

\item \textit{Resolve with} \(C_{\text{current}} = \text{Green}(\text{Dave}, \text{False})\):

  \[
  \begin{aligned}
  &\text{Green}(\text{Dave}, \text{False}) \\
  &\quad \text{and} \\
  &\text{Smart}(\text{Dave}, \text{False}) \lor \text{Green}(\text{Dave}, \text{True})
  \end{aligned}
  \]

\item \textit{Since} \(\text{Green}(\text{Dave}, \text{False})\) \textit{contradicts} \(\text{Green}(\text{Dave}, \text{True})\), \textit{the new resolved clause is}:

  \[
  C_{\text{resolved}} = \text{Smart}(\text{Dave}, \text{False})
  \]

\item \textit{Update \(C_\text{current}\) to \(C_\text{resolved}\)}:

  \[
  C_{\text{current}} = C_{\text{resolved}} = \text{Smart}(\text{Dave}, \text{False})
  \]

\end{itemize}

\medskip

\item \textbf{Third Resolution:}

\[
\text{Resolve}\left(C_{\text{current}} = \text{Smart}(\text{Dave}, \text{False}),\ C_{\text{complement}} = \text{Smart}(\text{Dave}, \text{True})\right)
\]

\begin{itemize}
\item \textit{Resolve} \(\text{Smart}(\text{Dave}, \text{False})\) \textit{with} \(\text{Smart}(\text{Dave}, \text{True})\):

  \[
  \begin{aligned}
  &\text{Smart}(\text{Dave}, \text{False}) \\
  &\quad \text{and} \\
  &\text{Smart}(\text{Dave}, \text{True})
  \end{aligned}
  \]

\item \textit{Since} \(\text{Smart}(\text{Dave}, \text{False})\) \textit{contradicts} \(\text{Smart}(\text{Dave}, \text{True})\), \textit{the final resolved clause is}:

  \[
  C_{\text{resolved}} = \text{Contradiction}
  \]

  Conclusion: \(D_{S_n}\) = \(P \vdash \neg S\)

\end{itemize}

\end{enumerate}

\end{tcolorbox}

We then start the second reasoning path from \(C_\text{current}\) = \(\neg S_n\).

\begin{tcolorbox}[title={Logical Resolution Steps}, colframe=blue!75!black, colback=white, fonttitle=\bfseries, breakable]

\textbf{Translated and Decomposed Premises \(P_n\):}

\begin{enumerate}
\item If someone is green then they are nice ::: \(\forall x \left( \text{Green}(x, \text{False}) \lor \text{Nice}(x, \text{True}) \right)\)
\item If someone is smart then they are green ::: \(\forall x \left( \text{Smart}(x, \text{False}) \lor \text{Green}(x, \text{True}) \right)\)
\item Dave is smart ::: \(\text{Smart}(\text{Dave}, \text{True})\)
\end{enumerate}

Question Statement \(S_n\): Dave is not nice ::: Nice(Dave, False)

\medskip

\(C_{\text{current}} = \neg S_n = \text{Nice}(\text{Dave}, \text{True})\)

\bigskip

\textbf{Resolution Steps:}

\begin{enumerate}
\item \textbf{First Resolution:}

No complementary clause was found from the given premises, thus we directly conclude "No contradiction found"

Conclusion: \(D_{\neg S_n}\) = \(P \not\vdash S\)

\medskip

\end{enumerate}

\end{tcolorbox}

Since we get: \(D_{S_n}\) = \(P \vdash \neg S\) and \(D_{\neg S_n}\) = \(P \not\vdash S\) from two reasoning paths correspondingly, according to Eq. (\ref{eq:logical_classification}), the final answer is False.

\clearpage

\section{Full Prompting of Each Module}

\subsection{ProntoQA}

\begin{tcolorbox}[breakable, title=Translation,colframe=blue!75!black, colback=white, fonttitle=\bfseries,before skip=1pt, after skip=1pt,fontupper=\linespread{0.8}\selectfont]
{\footnotesize

\textbf{Task Description:}

You are given a problem description and a question. The task is to:
\begin{enumerate}
    \item Define all the predicates in the problem.
    \item Parse the problem into logic rules based on the defined predicates.
    \item Write all the facts mentioned in the problem.
    \item Parse the question into the logic form.
\end{enumerate}

\textbf{Premises \(P\):}
\begin{itemize}
    \item Each jompus is fruity. Every jompus is a wumpus. Every wumpus is not transparent. Wumpuses are tumpuses. Tumpuses are mean. Tumpuses are vumpuses. Every vumpus is cold. Each vumpus is a yumpus. Yumpuses are orange. Yumpuses are numpuses. Numpuses are dull. Each numpus is a dumpus. Every dumpus is not shy. Impuses are shy. Dumpuses are rompuses. Each rompus is liquid. Rompuses are zumpuses. Alex is a tumpus.
\end{itemize}

\textbf{Statement \(S\):}
\begin{itemize}
    \item True or false: Alex is not shy.
\end{itemize}

\textbf{Facts (included in \(P_t\)):}
\begin{itemize}
    \item \(Tumpuses(Alex)\): Alex is a tumpus. 
    \item (... more facts ...)
\end{itemize}

\textbf{Rules (included in \(P_t\)):}
\begin{itemize}
    \item \(Jompus(x) \Rightarrow Fruity(x)\): Each jompus is fruity.
    \item (... more rules ...)
\end{itemize}

\textbf{Translated Query \(S_t\):}
\begin{itemize}
    \item \(Shy(Alex, False)\) ::: Alex is not shy
\end{itemize}

}
\end{tcolorbox}

\begin{tcolorbox}[breakable, title=Decomposition, colframe=blue!75!black, colback=white, fonttitle=\bfseries, before skip=1pt, after skip=1pt, fontupper=\linespread{0.8}\selectfont]
{\footnotesize

\textbf{Task Description:}

You are given a problem description and a question. The task is to:

\begin{enumerate}
    \item Given the premises and conjecture in logical form, decompose the logical statements using normalization and skolemization.
    \item \textbf{Normalization:} Convert each premise and conjecture into Prenex Normal Form (PNF), then into Conjunctive Normal Form (CNF).
    \item \textbf{Skolemization:} Eliminate existential quantifiers by introducing Skolem constants or functions.
\end{enumerate}

\textbf{Premises \(P_t\):}
\begin{itemize}
    \item \(Jompus(x, True) \rightarrow Shy(x, False)\)
    \item \(Jompus(x, True) \rightarrow Yumpus(x, True)\)
    \item (...more premises... )
\end{itemize}

\textbf{Query \(S_t\):}
\begin{itemize}
    \item \(Sour(Max, True)\)
\end{itemize}

\textbf{Decomposed Premises \(P_n\):}
\begin{itemize}
    \item 1. \( \neg Jompus(x, True) \lor Shy(x, False) \) 
    \item 2. \(( \neg Jompus(x, True) \lor Yumpus(x, True) \)
    \item (... additional decomposed premises ...)
\end{itemize}

\textbf{Query \(S_n\):}
\begin{itemize}
    \item \(Sour(Max, True)\)
\end{itemize}

}
\end{tcolorbox}

\begin{tcolorbox}[breakable, title=Resolve, colframe=blue!75!black, colback=white, fonttitle=\bfseries, before skip=1pt, after skip=1pt, fontupper=\linespread{0.8}\selectfont]
{\footnotesize

\textbf{Task Description:}

You are given a problem description and a question. The task is to:

\begin{enumerate}
    \item Check for Complementary/Contradictory Terms. Two terms are contradictory if they share the same predicate and arguments but differ in boolean value (True vs. False).
    \item If contradictory terms are found, apply the resolution rule: From \( (P(x, True) \lor Q(x, True)) \) and \( (P(x, False) \lor M(x, True)) \), derive \( Q(x, True) \lor M(x, True) \).
    \item If the resolution leads to an empty clause or direct contradiction, then output "Contradiction". Otherwise output the new clause after resolution.
\end{enumerate}

\textbf{Example:}
\textbf{Given Clauses (\(C_\text{current}\) and \(C_\text{complement}\))}
\begin{itemize}
    \item \(Difficult(Bradley, True) \lor Known(x,False) \)
    \item \(Difficult(x,False) \lor Embarrassed(x,True) \lor Colorful(x,False)\)
\end{itemize}

\textbf{Resolved \(C_\text{resolved}\):}
\begin{itemize}
    \item \( Known(x, False) \lor Embarrassed(Bradley, True) \lor Colorful(Bradley, False) \)
\end{itemize}

(...more examples...)

}
\end{tcolorbox}

\subsection{ProofWriter}

\begin{tcolorbox}[breakable, title=Translation,colframe=blue!75!black, colback=white, fonttitle=\bfseries,before skip=1pt, after skip=1pt,fontupper=\linespread{0.8}\selectfont]
{\footnotesize

\textbf{Task Description:}

You are given a problem description and a question. The task is to:
\begin{enumerate}
    \item Define all the predicates in the problem.
    \item Parse the problem into logic rules based on the defined predicates.
    \item Write all the facts mentioned in the problem.
    \item Parse the question into the logic form.
\end{enumerate}

\textbf{Premises \(P\):}
\begin{itemize}
    \item Anne is quiet. Erin is furry. Erin is green. Fiona is furry. Fiona is quiet. Fiona is red. Fiona is rough. Fiona is white. Harry is furry. Harry is quiet. Harry is white. Young people are furry. If Anne is quiet then Anne is red. Young, green people are rough. If someone is green then they are white. If someone is furry and quiet then they are white. If someone is young and white then they are rough. All red people are young.
\end{itemize}

\textbf{Statement \(S\):}
\begin{itemize}
    \item Is the following statement true, false, or unknown? Anne is white.
\end{itemize}

\textbf{Facts (included in \(P_t\)):}
\begin{itemize}
    \item \(Quite(Anne, True)\): Anne is quiet.
    \item (... More facts ...)
\end{itemize}

\textbf{Rules (included in \(P_t\)):}
\begin{itemize}
    \item \(Young(x, True) \Rightarrow Furry(x, True)\): Young people are furry.
    \item (... More rules ...)
\end{itemize}

\textbf{Query \(S_t\):}
\begin{itemize}
    \item White(Anne, True) ::: Anne is white.
\end{itemize}

}
\end{tcolorbox}

\begin{tcolorbox}[breakable, title=Decomposition, colframe=blue!75!black, colback=white, fonttitle=\bfseries, before skip=1pt, after skip=1pt, fontupper=\linespread{0.8}\selectfont]
{\footnotesize

\textbf{Task Description:}

You are given a problem description and a question. The task is to:

\begin{enumerate}
    \item Given the premises and conjecture in logical form, decompose the logical statements using normalization and skolemization.
    \item \textbf{Normalization:} Convert each premise and conjecture into Prenex Normal Form (PNF), then into Conjunctive Normal Form (CNF).
    \item \textbf{Skolemization:} Eliminate existential quantifiers by introducing Skolem constants or functions.
\end{enumerate}

\textbf{Premises \(P_t\):}
\begin{itemize}
    \item \(Quite(Anne, True)\): Anne is quiet.
    \item \(Young(x, True) \Rightarrow Furry(x, True)\): Young people are furry.
    \item (...more premises... )
\end{itemize}

\textbf{Query \(S_t\):}
\begin{itemize}
    \item White(Anne, True) ::: Anne is white.
\end{itemize}

\textbf{Decomposed Premises \(P_n\):}
\begin{itemize}
    \item 1. \(Quite(Anne, True)\)
    \item 2. \(Young(x, False) \lor Furry(x, True)\)
    \item (... additional decomposed premises ...)
\end{itemize}

\textbf{Query \(S_n\):}
\begin{itemize}
    \item \(White(Anne, True)\)
\end{itemize}

}
\end{tcolorbox}

\begin{tcolorbox}[breakable, title=Resolve, colframe=blue!75!black, colback=white, fonttitle=\bfseries, before skip=1pt, after skip=1pt, fontupper=\linespread{0.8}\selectfont]
{\footnotesize

\textbf{Task Description:}

You are given a problem description and a question. The task is to:

\begin{enumerate}
    \item Check for Complementary/Contradictory Terms. Two terms are contradictory if they share the same predicate and arguments but differ in boolean value (True vs. False).
    \item If contradictory terms are found, apply the resolution rule: From \( (P(x, True) \lor Q(x, True)) \) and \( (P(x, False) \lor M(x, True)) \), derive \( Q(x, True) \lor M(x, True) \).
    \item If the resolution leads to an empty clause or direct contradiction, then output "Contradiction". Otherwise output the new clause after resolution.
\end{enumerate}

\textbf{Example:}
\textbf{Given Clauses (\(C_\text{current}\) and \(C_\text{complement}\))}
\begin{itemize}
    \item \(Difficult(Bradley, True) \lor Known(x,False) \)
    \item \(Difficult(x,False) \lor Embarrassed(x,True) \lor Colorful(x,False)\)
\end{itemize}

\textbf{Resolved \(C_\text{resolved}\):}
\begin{itemize}
    \item \( Known(x, False) \lor Embarrassed(Bradley, True) \lor Colorful(Bradley, False) \)
\end{itemize}

(...more examples...)

}
\end{tcolorbox}

\subsection{LogicNLI}

\begin{tcolorbox}[breakable, title=Translation,colframe=blue!75!black, colback=white, fonttitle=\bfseries,before skip=1pt, after skip=1pt,fontupper=\linespread{0.8}\selectfont]
{\footnotesize

\textbf{Task Description:}

You are given a problem description and a question. The task is to:
\begin{enumerate}
    \item Define all the predicates in the problem.
    \item Parse the problem into logic rules based on the defined predicates.
    \item Write all the facts mentioned in the problem.
    \item Parse the question into the logic form.
    \item Please make sure to differentiate 'or' and 'either…or…'. For 'or', you should translate it with the inclusive 'or' ($\lor$) operator. For 'either…or…', you should translate it with the 'exclusive or' ($\oplus$) operator.
    \item Please be careful when translating clauses with words "equivalent", "vice versa" and "if and only if". Make sure you use the biconditional "$\leftrightarrow$" in those translations.
\end{enumerate}

\textbf{Premises \(P\):}
\begin{itemize}
    \item Medwin is doubtful. Roberto is not bitter. Roberto is not grieving. If someone is not bitter, then he is not grieving. Medwin being not sociable implies that Medwin is not pure. If there is someone who is either not pure or not doubtful, then Lynda is not grieving.
\end{itemize}

\textbf{Statement \(S\):}
\begin{itemize}
    \item Bernard is not bitter.
\end{itemize}

\textbf{Facts (included in \(P_t\)):}
\begin{itemize}
    \item \(Doubtful(Medwin, True)\) ::: Medwin is doubtful.
    \item (... More facts ...)
\end{itemize}

\textbf{Rules (included in \(P_t\)):}
\begin{itemize}
    \item \(Bitter(x, False) \Rightarrow Grieving(x, False)\): If someone is not bitter, then he is not grieving.
    \item (... More rules ...)
\end{itemize}

\textbf{Query \(S_t\):}
\begin{itemize}
    \item Bitter(Bernard, False) ::: Bernard is not bitter.
\end{itemize}

}
\end{tcolorbox}

\begin{tcolorbox}[breakable, title=Decomposition, colframe=blue!75!black, colback=white, fonttitle=\bfseries, before skip=1pt, after skip=1pt, fontupper=\linespread{0.8}\selectfont]
{\footnotesize

\textbf{Task Description:}

You are given a problem description and a question. The task is to:

\begin{enumerate}
    \item Given the premises and conjecture in logical form, decompose the logical statements using normalization and skolemization.
    \item \textbf{Normalization:} Convert each premise and conjecture into Prenex Normal Form (PNF), then into Conjunctive Normal Form (CNF).
    \item \textbf{Skolemization:} Eliminate existential quantifiers by introducing Skolem constants or functions.
\end{enumerate}

\textbf{Premises \(P_t\):}
\begin{itemize}
    \item \(Doubtful(Medwin, True)\) ::: Medwin is doubtful.
    \item \(Bitter(x, False) \Rightarrow Grieving(x, False)\): If someone is not bitter, then he is not grieving.
    \item (...more premises... )
\end{itemize}

\textbf{Query \(S_t\):}
\begin{itemize}
    \item \(Bitter(Bernard, False)\) ::: Bernard is not bitter.
\end{itemize}

\textbf{Decomposed Premises \(P_n\):}
\begin{itemize}
    \item 1. \(Doubtful(Medwin, True)\)
    \item 2. \(Bitter(x, True) \lor Grieving(x, False)\)
    \item (... additional decomposed premises ...)
\end{itemize}

\textbf{Query \(S_n\):}
\begin{itemize}
    \item \(Bitter(Bernard, False)\)
\end{itemize}

}
\end{tcolorbox}

\begin{tcolorbox}[breakable, title=Resolve, colframe=blue!75!black, colback=white, fonttitle=\bfseries, before skip=1pt, after skip=1pt, fontupper=\linespread{0.8}\selectfont]
{\footnotesize

\textbf{Task Description:}

You are given a problem description and a question. The task is to:

\begin{enumerate}
    \item Check for Complementary/Contradictory Terms. Two terms are contradictory if they share the same predicate and arguments but differ in boolean value (True vs. False).
    \item If contradictory terms are found, apply the resolution rule: From \( (P(x, True) \lor Q(x, True)) \) and \( (P(x, False) \lor M(x, True)) \), derive \( Q(x, True) \lor M(x, True) \).
    \item If the resolution leads to an empty clause or direct contradiction, then output "Contradiction". Otherwise output the new clause after resolution.
\end{enumerate}

\textbf{Example:}
\textbf{Given Clauses (\(C_\text{current}\) and \(C_\text{complement}\))}
\begin{itemize}
    \item \(Difficult(Bradley, True) \lor Known(x,False) \)
    \item \(Difficult(x,False) \lor Embarrassed(x,True) \lor Colorful(x,False)\)
\end{itemize}

\textbf{Resolved \(C_\text{resolved}\):}
\begin{itemize}
    \item \( Known(x, False) \lor Embarrassed(Bradley, True) \lor Colorful(Bradley, False) \)
\end{itemize}

(...more examples...)

}
\end{tcolorbox}

\end{document}